\documentclass[preprint]{article}

\usepackage[preprint]{neurips_2026}

\usepackage[utf8]{inputenc} 
\usepackage[T1]{fontenc}    
\usepackage{hyperref}       
\usepackage{url}            
\usepackage{booktabs}       
\usepackage{amsfonts}       
\usepackage{nicefrac}       
\usepackage{microtype}      
\usepackage{xcolor}         
\usepackage{enumitem}       
\usepackage{algorithm}      
\usepackage{algpseudocode}  
\usepackage{graphicx}       
\usepackage{amsfonts}   
\usepackage{amsmath}

\title{Spectral Scaling Laws of Muon}

\author{
  Gagik Magakyan\thanks{Correspondence to: \texttt{gagmag@mit.edu}.} \\
  MIT
  \And
  Pablo Parrilo \\
  MIT
  \And
  Asuman Ozdaglar \\
  MIT
}

\begin{document}

\maketitle

\begin{abstract}
    Orthonormalized update rules have rapidly become a leading
    choice of optimizer for training large language models, with
    recent open-source state-of-the-art models adopting Muon. To keep
    these updates tractable, Muon performs the orthonormalization
    with the Newton--Schulz (NS) iteration. Since NS is only
    approximate, directions with small singular values fail to be
    orthonormalized. In Muon, NS is applied to the momentum matrix
    at every step, yet little is known about how the singular value
    spectrum of these momentum matrices behaves during training, or
    how that behavior changes with model size. We present the first
    systematic study of this question. Tracking singular value
    quantiles of the momentum buffer across layers in models ranging
    from 77M to 2.8B parameters, we observe a consistent picture:
    after a short burn-in, the quantiles stabilize at a value
    determined by the layer type and model size. These stabilization
    values follow remarkably clean power laws in model size, with
    layer-dependent exponents. Layers up to mid-late depth scale very
    mildly with model size $M$ (around $M^{-0.25}$), so the standard
    5-step NS configuration used at academic scale will continue to
    orthonormalize them at much larger scales. Some of the late
    layers, however, scale much more aggressively (up to $M^{-0.96}$)
    and will fall into the NS failure regime at frontier scale unless
    one uses more NS iterations or better-tuned coefficients. NS
    iterations are computationally expensive at scale; our laws give
    practitioners a principled, layer-aware recipe for choosing the
    minimum NS configuration that still orthonormalizes the
    directions that matter --- avoiding unnecessary computation
    without sacrificing update quality.
\end{abstract}

\section{Introduction}  

\label{sec:introduction}

Pre-training large language models (LLMs) is a costly process that consumes
millions of GPU hours, making the choice of optimizer a central design
decision: even modest gains in optimizer efficiency translate into substantial
savings. AdamW \citep{adamw_2019,adam_2015} has long been the standard optimizer for training
LLMs \citep{deepseekv3_2024,llama3_2024,olmo2_2025}.
More recently, orthonormalized-update optimizers such as Muon \citep{jordan2024muon, bernstein2025modular} have begun to take its place, providing more stable training and better hyperparameter transfer\citep{pethick2025training} across scales. At larger scale, \citet{liu2025muon} show that Muon achieves twice the compute efficiency of AdamW. Notably, the recent state-of-the-art models Kimi-K2, GLM-5, and DeepSeek-V4 \citep{kimi2026k25, glm5_2026, deepseekv4} were all trained with Muon.

Muon performs approximate orthonormalization of the momentum matrices using
the Newton--Schulz (NS) iteration \citep{higham2008functions, kovarik1970iterative, bjorck1971iterative}, which repeatedly applies an odd polynomial to push each singular value toward $1$.
Since NS is only approximate, directions whose singular values are too small
fail to be properly orthonormalized (see \autoref{fig:standard_ns_map}).
Whether a given NS configuration is accurate enough therefore depends on where the
momentum singular values actually reside during training: if they are large,
even a cheap NS configuration orthonormalizes them correctly; if they are small,
more accurate one is required.
The academic community typically uses the $5$-polynomial NS coefficients introduced by \citet{cesista2025muonoptcoeffs}, which were popularized by the NanoGPT speedrun \citep{modded_nanogpt_2024}. The recent frontier-scale DeepSeek-V4~\citep{deepseekv4} uses a more accurate composition of $10$  polynomials.
In both cases the same configuration is applied uniformly across all
layers. Since each NS step carries a non-trivial cost at scale
\citep{essentialai2025,ahn2025dion},
a natural question is whether $5$ polynomials already suffice at scale, or whether $10$ are needed --- and crucially, whether the answer is the same for every layer.

\begin{figure}[h]
    \centering
    \includegraphics[width=1.0\textwidth]{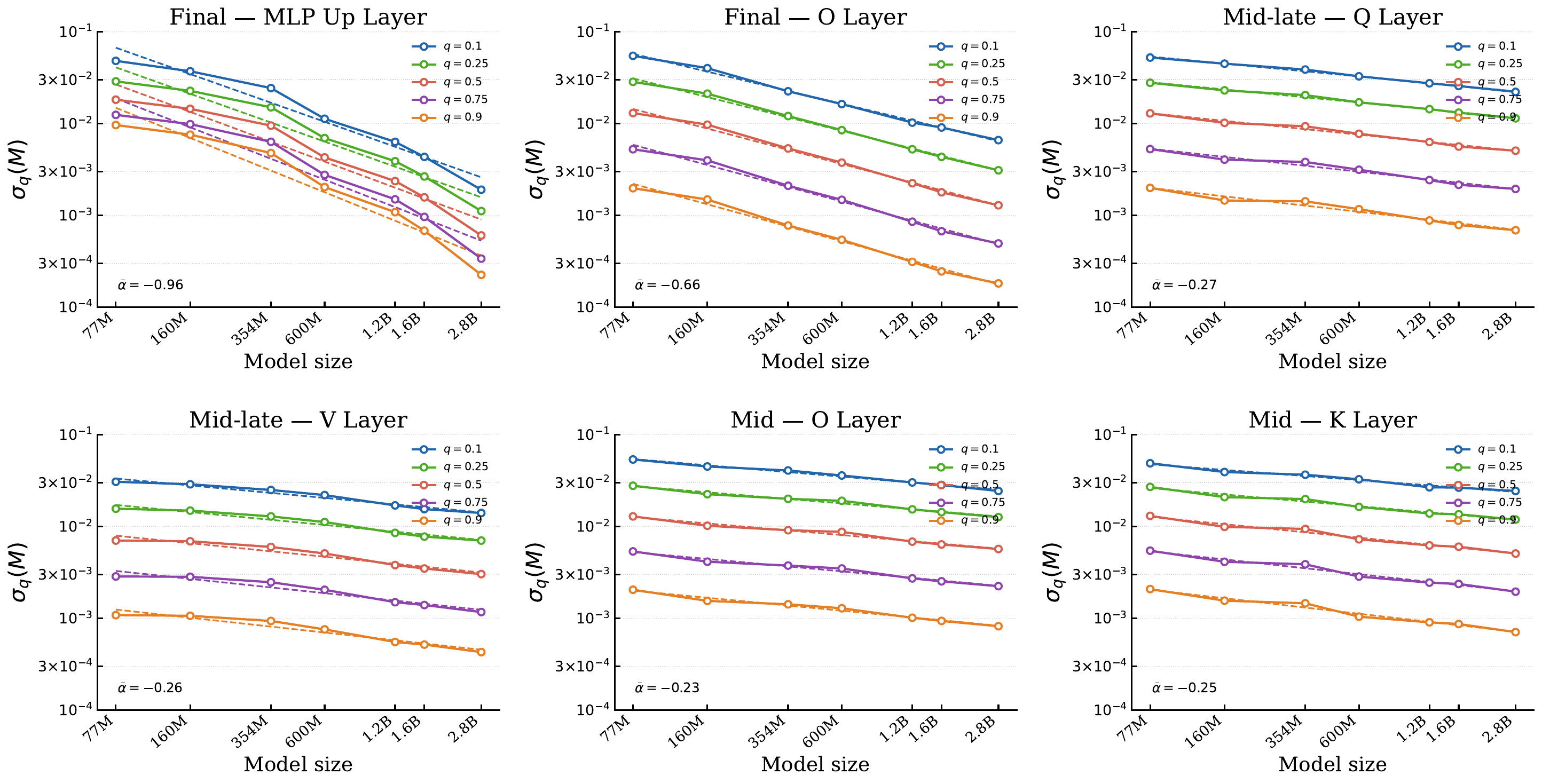}
    \caption{Scaling laws for the stabilization values of the singular value
    quantiles of normalized momentum matrices of different depth layers and model sizes.}
    \label{fig:scaling_laws}
\end{figure}

To answer this, we conduct the first systematic study of how the
singular values of Muon's momentum matrices evolve during training,
tracking quantiles at multiple depths in  GPT-2-style models ranging from 77M to
2.8B parameters. A consistent picture emerges across all layers and
model sizes: after a short burn-in period, the singular value
quantiles stabilize around a value that depends on the layer type
and decreases with model size. Fitting power laws to these
stabilization values reveals a remarkably clean log-log linear
relationship with layer-dependent exponents (see
\autoref{fig:scaling_laws}). This lets us extrapolate, at each depth,
how accurate NS must be to orthonormalize enough directions to preserve update quality
at frontier scale.

The exponents vary substantially across depth. Layers up to mid-late depth scale very mildly with model size (around $M^{-0.25}$); the NS approximation used in the NanoGPT experiments remains accurate enough for them at much larger scales. Some of the late layers, however, scale much more aggressively (up to $M^{-0.96}$) and will fall into the NS failure regime at frontier scale unless one uses more NS iterations or better-tuned coefficients.

Together, these findings let practitioners choose, layer by layer,
the cheapest NS configuration that still orthonormalizes the
directions that matter at any target scale.
Our main contributions are:
\begin{itemize}
    \item The first systematic study of the singular value spectrum
    of Muon's momentum buffer across layers and model sizes
    (77M--2.8B).
    \item Spectral power laws relating stabilization values to model
    size, with layer-dependent exponents.
    \item A practical recipe for selecting layer-specific NS
    configurations at frontier scale, derived directly from the
    fitted laws.
\end{itemize}

\subsection{Related work}

Several works address the cost of running Muon at scale when weight
matrices are sharded across devices. \citet{ahn2025dion, ahn2025dion2}
propose Dion, a distributed optimizer that achieves communication-efficient
 orthonormalized updates via low-rank approximations, while
  \citet{khaled2026muonbp} instead apply NS independently on each shard with periodic global synchronization for training stability.

A parallel line of work explores matrix-preconditioned optimizers for
deep learning, including Shampoo \citep{gupta2018shampoo}, SOAP
\citep{vyas2024soap}, and COSMOS \citep{liu2026cosmos}.
\citet{anil2020scalable} and \citet{shi2023distributed} make Shampoo
practical at scale, though it requires heuristics such as learning
rate grafting~\citep{agarwal2020disentangling} to match Adam in
practice and, to our knowledge, has not yet been adopted at frontier
scale. \citet{eschenhagen2025purifying} mitigates some of these
heuristics by adaptively updating the preconditioner. Closer to Muon,
\citet{li2025normuon} augments orthonormalized updates with
Adam-style second moments, adding adaptive per-coordinate scaling on
top of Muon's spectral-norm step. \citet{wen2025fantastic} benchmark
many of these optimizers across model sizes and data-to-model ratios.

Scaling laws were pioneered by \citet{kaplan2020scaling}, who showed
that language model loss follows clean power laws in parameters,
training tokens, and compute. \citet{hoffmann2022chinchilla} refined
these relationships into compute-optimal token-to-parameter ratios,
establishing that prior large models were significantly undertrained.
A complementary direction asks what optimizer hyperparameters scale
predictably with model size. \citet{yang2022tensorprograms} show that
optimal learning rates transfer zero-shot across scales under the
$\mu$P parameterization, demonstrating that optimizer hyperparameters
obey their own scaling structure. We contribute to these lines of work: we show that, after a short burn-in, the singular value
quantiles of Muon's momentum buffers stabilize at values that follow
power laws in model size, with layer-dependent exponents.

\section{Background: Muon and Newton-Schulz }

This section establishes the background for the rest of the paper.
We first describe the Muon optimizer (\autoref{sec:muon}) and review
the Newton-Schulz iteration it uses for approximate
orthonormalization (\autoref{sec:ns}), highlighting its key
limitation: directions with sufficiently small singular values fail
to be orthonormalized. We then run a controlled experiment
(\autoref{sec:how_much_normalization_is_needed}) to determine which
fraction of singular directions must be orthonormalized for Muon to
retain its benefits, which fixes the quantile range we track for the
rest of the paper.

\subsection{The Muon Optimizer}
\label{sec:muon}

Muon~\citep{jordan2024muon,bernstein2025modular} replaces the raw
gradient update on 2D weight matrices with an \emph{orthonormalized}
update.At each step Muon maintains a momentum
buffer $M_t$ and applies a Newton-Schulz (NS) iteration to
approximately orthonormalize it before stepping the parameters
(See Algorithm \ref{alg:muon}).

\begin{algorithm}[h]
\caption{Muon Optimizer}
\label{alg:muon}
\begin{algorithmic}[1]
\Require Learning rate $\eta$, momentum coefficient $\mu$, initial parameters $\Theta_0$
\State Initialize momentum buffer $M_0 \leftarrow 0$
\For{$t = 0, 1, 2, \ldots$}
    \State Compute gradient $G_t = \nabla_{\Theta} \mathcal{L}(\Theta_t)$
    \State Update momentum: $M_{t+1} \leftarrow \mu \cdot M_t + G_t$
    \State Orthonormalize: $O_{t+1} \leftarrow \mathrm{NS}(M_{t+1})$
    \State Update parameters: $\Theta_{t+1} \leftarrow \Theta_t - \eta \cdot O_{t+1}$
\EndFor
\end{algorithmic}
\end{algorithm}

Here $\mathrm{NS}(\cdot)$ denotes the Newton-Schulz iteration 
described in \autoref{sec:ns}. 
Muon is motivated by steepest descent under 
the spectral norm \citep{bernstein2025modular}, and has been shown to double the 
compute efficiency compared to AdamW
on language model training tasks in scale\citep{liu2025muon}.

\subsection{Newton-Schulz Iteration for Approximate Orthonormalization}
\label{sec:ns}

We now focus our attention on the problem of approximate orthonormalization.
 Exact orthonormalization is expensive for large matrices, so practical 
implementations use fast iterative approximations.

The standard approximation in this line of work is the Newton-Schulz (NS) 
iteration. Let $A$ be a momentum matrix to be orthonormalized. NS first normalizes
\[
\widetilde{A}_0 = \frac{A}{\|A\|_F},
\]
to transform all singular values to $[0,1]$ interval and then applies a sequence of 
odd-degree polynomials $p_0, \dots, p_{n-1}$. In practice, each $p_k$ is taken to be a degree-5 odd polynomial of the
form
\begin{align}
\widetilde{A}_{k+1} = p_k(\widetilde{A}_k) = 
a_k\widetilde{A}_k + b_k \left(\widetilde{A}_k \widetilde{A}_k^{\top} \right)\widetilde{A}_k
+ c_k \left(\widetilde{A}_k \widetilde{A}_k^{\top} \right)^2 \widetilde{A}_k, \quad k = 0,1,\ldots,n-1,
\label{eq:ns_iteration}
\end{align}
Recall that for any matrix with SVD $A = U S V^\top$,
an odd polynomial satisfies
\[
p(A) = U\, p(S)\, V^\top,
\]                                                                                                                  
where $p(S)$ applies $p$ elementwise to the diagonal of $S$. This means the singular vectors
are \emph{exactly preserved} at every step, and only the singular values are modified.
Unrolling \autoref{eq:ns_iteration} $n$ times, the final result is
\[
\widetilde{A}_n = U\, \underbrace{(p_n \circ \cdots \circ p_1)(S)}_{=:\, f(S)}\, V^\top,
\]
Thus, the NS procedure reduces to a one-dimensional problem: find a scalar composition
$f = p_n \circ \cdots \circ p_1$ such that $f(\sigma) \approx 1$ for every singular value
$\sigma \in [0, 1]$. In other words, $f$ should approximate the sign function on $(0,1]$,
pushing every singular value toward $1$ regardless of where it starts.
When this condition holds, $\widetilde{A}_n \approx UV^\top$, recovering the orthonormal
factor in the polar decomposition of $A$. However, since each $p_i$ is an odd polynomial,
$f(0) = 0$ for any choice of polynomials, so $f$ cannot approximate the sign
function in a neighborhood of zero---a fundamental limitation of the NS family.

\begin{figure}[h]
    \centering
    \includegraphics[width=0.8\textwidth]{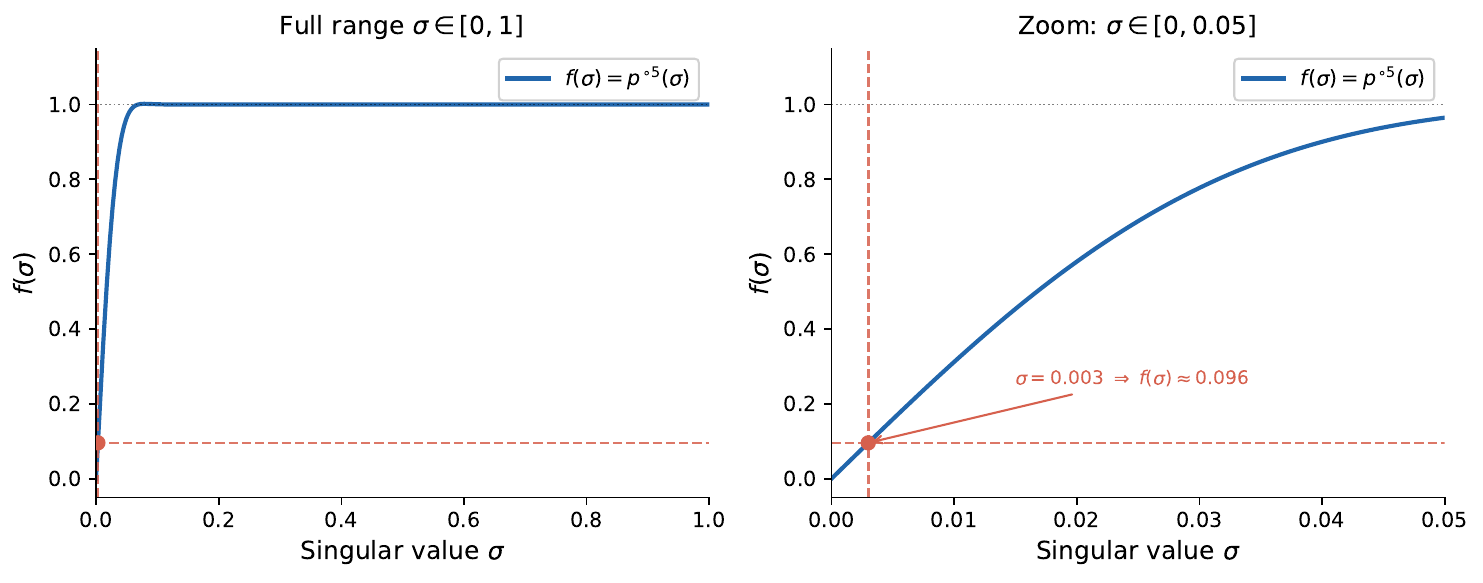}
    \caption{The NS map $f(\sigma) = p^{\circ 5}(\sigma)$ for the canonical polynomial $p(x) = 2x - 1.5x^3 + 0.5x^5$.
    The left plot shows the full range $\sigma \in [0,1]$. The right plot shows a zoom-in view of the region $\sigma \in [0, 0.05]$.}
    \label{fig:standard_ns_map}
\end{figure}

As a concrete example, consider the canonical polynomial used for NS and in 
the introduction of
Muon \citep{jordan2024muon}:
\[
    p(x) = 2x - 1.5x^3 + 0.5x^5,
\]
applied $n=5$ times, i.e.\ $f = p^{\circ 5}$.
\autoref{fig:standard_ns_map} plots $f(\sigma)$ as a function of $\sigma \in [0,1]$.
One can see that the approximation is accurate for $\sigma > 0.05$, pushing those values close to $1$.
However, the composition is approximately linear near the origin: for $\sigma \leq 0.003$
one can verify numerically that $f(\sigma) \leq 0.1$. In other words, any direction
whose singular value falls below roughly $0.003$ will remain essentially
\emph{unorthonormalized} after five NS steps — its effective contribution to the update
is suppressed by a factor of $10\times$ or more relative to a direction with large singular value.

In practice, different implementations use different NS configurations. The NanoGPT speedrun
\citep{modded_nanogpt_2024} uses optimized 5-step 
polynomials (see \autoref{fig:nanogpt_ns_map}), 
while DeepSeek-V4 \citep{deepseekv4} employs a 
more accurate 10-step composition 
(see \autoref{fig:deepseek_v4_ns_map}). Since each NS step carries
a non-trivial cost at scale \citep{essentialai2025, ahn2025dion}, 
a natural question is whether the additional steps are necessary 
to maintain update quality. To answer this, one must understand 
how the singular values of the momentum matrices actually behave 
during training — which we study systematically in 
\autoref{sec:quantiles}.

To understand which quantiles of the momentum spectrum are 
practically relevant, we run a controlled experiment with 
rank-$p$ orthonormal updates — using only the top $p$ 
fraction of singular directions — and measure how closely
 they track full Muon's performance. This determines 
 which quantiles to focus on in the sections that follow.

\subsection{How much orthonormalization is needed?}
\label{sec:how_much_normalization_is_needed}

We now investigate how many singular directions must be orthonormalized to retain the
benefits of Muon. To this end, we introduce \emph{rank-$p$ orthonormal updates}: given
the SVD $M = USV^\top \in \mathbb{R}^{m \times n}$ of the momentum matrix, the update
direction is formed using only the top-$k$ singular vectors,
\[
    O = U_{:,\,1:k}\,V_{:,\,1:k}^\top, \qquad
    k = \left\lfloor \min(m, n) \cdot p \right\rfloor,
\]
where $p \in \{0.1,\,0.25,\,0.5,\,0.9\}$ denotes the fraction of
singular directions retained. We pretrain GPT-2-style models with
77M, 160M, and 354M parameters (\autoref{tab:model_configs}) across
this range of $p$ values, each for a Chinchilla-optimal number of
tokens~\citep{hoffmann2022chinchilla}. For more details see \autoref{sec:pre_training_details}.

\begin{figure}[h]
    \centering
    \includegraphics[width=1.0\textwidth]{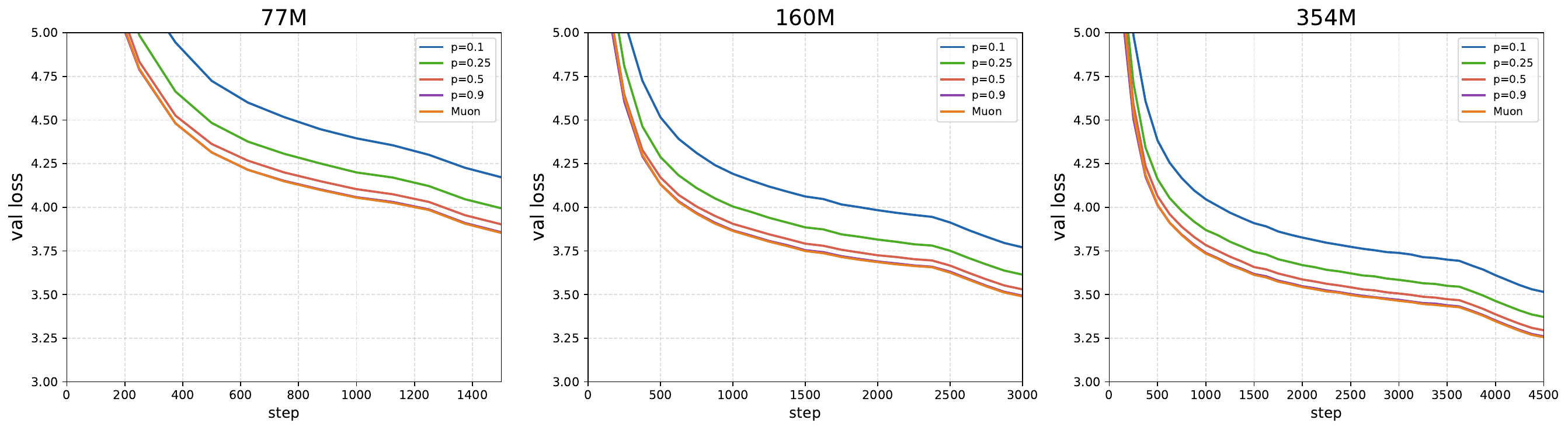}
    \caption{Pre-training models of sizes 77M, 160M, and 354M parameters with rank-$p$ 
    orthonormal updates. }
    \label{fig:ablation_directions}
\end{figure}

\autoref{fig:ablation_directions} reveals a monotonic
degradation as $p$ decreases: $p=0.9$ is essentially indistinguishable
from full Muon, and even $p=0.5$ incurs only a minor performance gap.
Quantifying these gaps more precisely
(\autoref{fig:ablation_directions_matching_25},
\autoref{fig:ablation_directions_matching_10}), $p=0.25$ updates are
around $10\text{--}20\%$ less token-efficient than full Muon which is a gap
that may be acceptable in practice. $p=0.1$, in contrast, is around
$50\%$ less efficient and impractical.

\citet{ahn2025dion} run a similar low-rank ablation for the Dion
optimizer (their Figure~2) and observe that the performance gap
relative to full Dion narrows with model scale. Their setting
differs from ours in one important way: Dion uses \emph{error
feedback}, accumulating the residual of the low-rank approximation
back into the momentum buffer, which compensates for the
discarded information. Our rank-$p$ updates have no such
compensation, so we do not expect the same narrowing trend.

\begin{figure}[h]
    \centering
    \includegraphics[width=0.9\textwidth]{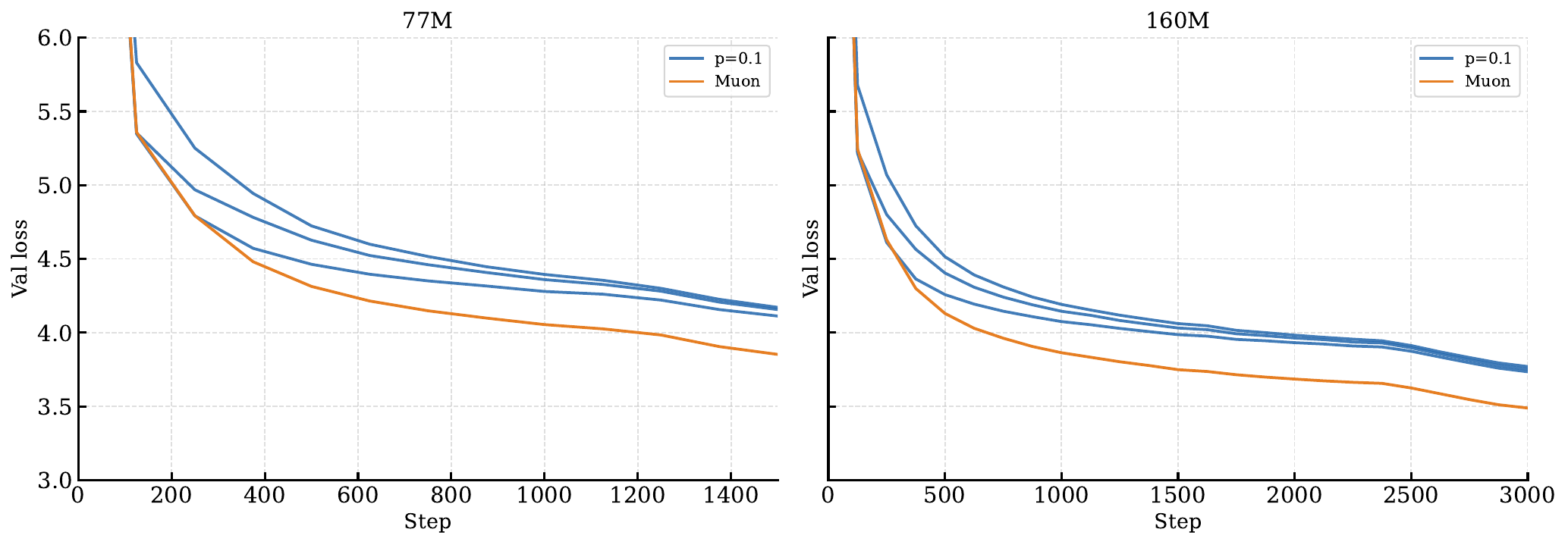}
    \caption{We compare training with low-rank $r=0.1$ updates from 
    scratch against first running full Muon for 125 (or 250) steps
     and then switching to low-rank updates. In all cases the gap 
     relative to full Muon remains large, indicating that the 
     performance difference is not simply due to Muon's faster
      convergence at the start of training.}
    \label{fig:start_svd_ablation}
\end{figure}

A potential concern is that the validation curves in
\autoref{fig:ablation_directions} run parallel after the initial
phase, suggesting the gap may stem from Muon's faster convergence
early in training rather than from a fundamental advantage of full
orthonormalization. To check this, we pre-train 77M and 160M models
with full Muon for 125 and 250 steps respectively — well into the
regime where its advantage over $p=0.1$ has already opened up — and
then switch to $p=0.1$ updates. As shown in
\autoref{fig:start_svd_ablation}, the gap relative to full Muon
remains large in both cases, confirming that the gap is not an
artifact of early-training dynamics.

\paragraph{Takeaway.} Orthonormalizing roughly the top half of
singular directions is enough to recover (or nearly recover) full
Muon, but orthonormalizing only the top $10\%$ is not. To
understand which NS approximations are needed to orthonormalize this
range of directions, we now turn to the singular value spectrum of
the momentum matrices.

\section{Spectral Dynamics of the Momentum Buffer}
\label{sec:quantiles}

In this section we track the quantiles of normalized singular values of the 
momentum matrices and observe how they evolve during the training. Before
proceeding we provide necessary notation and setup.

\paragraph{Notation.}
Let $A \in \mathbb{R}^{m \times n}$ be a matrix with singular values
sorted in descending order $\sigma_1(A) \geq \sigma_2(A) \geq \cdots
\geq \sigma_r(A)$, where $r = \min(m,n)$. For $q \in (0, 1]$ we
define the \emph{$q$-quantile singular value}
\[
    \sigma_q(A) := \sigma_{\lceil q \cdot r \rceil}(A),
\]
so that $\sigma_{0.5}(A)$ is the median (roughly half of singular
values are larger) and $\sigma_{1.0}(A) = \sigma_r(A)$ is the
smallest. Note that under this convention $\sigma_{0.1}(A)$ is a
\emph{large} singular value (only $\sim\!10\%$ are larger) and
$\sigma_{0.9}(A)$ is a \emph{small} one. We track these quantiles
for $A = M^{(t)} / \|M^{(t)}\|_F$, the Frobenius-normalized momentum
matrix of a given layer at training step $t$. This is exactly the
input that NS sees as $\widetilde{A}_0$ (\autoref{sec:ns}), so the
tracked quantiles are directly comparable to NS's failure threshold.
We use them to understand how the singular value spectrum evolves
over training and to quantify the fraction of directions that a
given NS configuration fails to orthonormalize.

\paragraph{Setup.}
We pretrain a suite of GPT-2-style language models ranging from 77M
to 2.8B parameters with Muon; configurations are detailed in
\autoref{tab:model_configs}. Each model is trained for the
Chinchilla-optimal number of tokens~\citep{hoffmann2022chinchilla}.

\begin{table}[h]
\centering
\caption{Model configurations used in our experiments.}
\label{tab:model_configs}
\begin{tabular}{lrrrrr}
\toprule
\textbf{Model} & \textbf{Params} & \textbf{Model Dim} & \textbf{\# Layers} & \textbf{\# Heads} & \textbf{Seq Len} \\
\midrule
GPT2-77M   &  77M  &  512 &  8 &  8 &  512 \\
GPT2-160M  & 160M  &  768 & 12 & 12 &  512 \\
GPT2-354M  & 354M  & 1024 & 20 & 16 &  512 \\
GPT2-600M  & 600M  & 1280 & 24 & 20 & 512 \\
GPT2-1.2B  & 1.2B  & 1792 & 26 & 28 & 1024 \\
GPT2-1.6B  & 1.6B  & 2048 & 28 & 32 & 1024 \\
GPT2-2.8B  & 2.8B  & 2560 & 32 & 40 & 1024 \\
\bottomrule
\end{tabular}
\end{table}

Since models vary in depth across configurations, we select four \emph{relative depth
checkpoints} to ensure comparability across model sizes. Concretely, for a model with
$N$ transformer layers we monitor layers
$
    \left\lfloor \frac{N}{4} \right\rfloor,
    \left\lfloor \frac{2N}{4} \right\rfloor,
    \left\lfloor \frac{3N}{4} \right\rfloor
    N,
$
corresponding to the \emph{mid-early}, \emph{mid}, \emph{mid-late},
and \emph{final} layers of the network. Within each selected layer we
track all six momentum matrices in the block --- the four attention
projections $Q$, $K$, $V$, $O$ and the two MLP projections. This
gives $4 \times 6 = 24$ momentum buffers per model. For each, we
record the singular value quantiles $\sigma_q(M^{(t)})$ for $q \in
\{0.1, 0.25, 0.5, 0.75, 0.9\}$ at every training step $t$.

\subsection{Stabilization of Singular Value Quantiles}
\begin{figure}[h]
    \centering
    \includegraphics[width=1.0\textwidth]{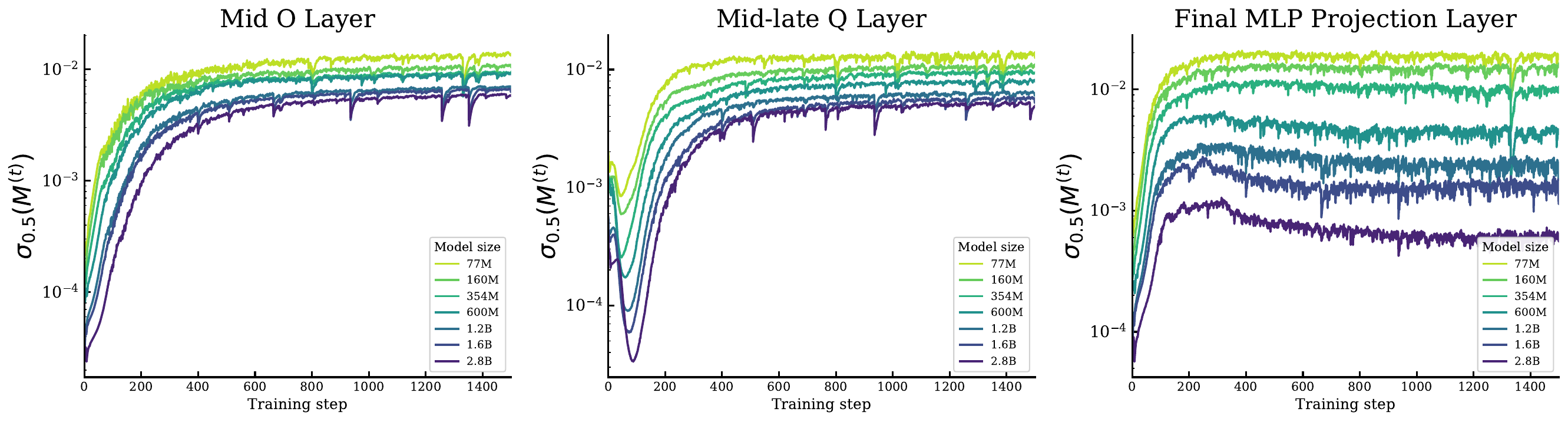}
    \caption{Quantile evolution for the 50\% quantile of the normalized singular values 
    for 3 fixed layer types and model sizes.}
    \label{fig:quantile_evolution}
\end{figure}

We now investigate how the tracked quantiles evolve during training.
\autoref{fig:quantile_evolution} plots $\sigma_{0.5}(M^{(t)})$ for
three layer types across different model sizes over the first $1500$
training steps. A consistent phenomenon emerges across model sizes
and layer types: after a short transient phase, the quantiles
stabilize at values that persist for the remainder of training. The
same pattern holds for all tracked layer types and quantiles
(\autoref{fig:quantile_evolution_25},
\autoref{fig:quantile_evolution_10}).

Notably, the shape of the transient differs by matrix type. For $Q$
and $K$ matrices, the quantiles exhibit a sharp decrease followed by
a recovery before stabilizing, whereas for $V$, $O$, and MLP matrices
the quantiles increase monotonically from the start before
stabilizing. We also observe that the stabilized values decrease
monotonically with model size, suggesting that as model size
increases a fixed NS configuration will fail to orthonormalize an
increasing fraction of directions --- a hypothesis we make
quantitative in \autoref{sec:scaling_laws}.

\subsection{Stabilization of the Full Spectrum}

\begin{figure}[h]
    \centering
    \includegraphics[width=0.9\textwidth]{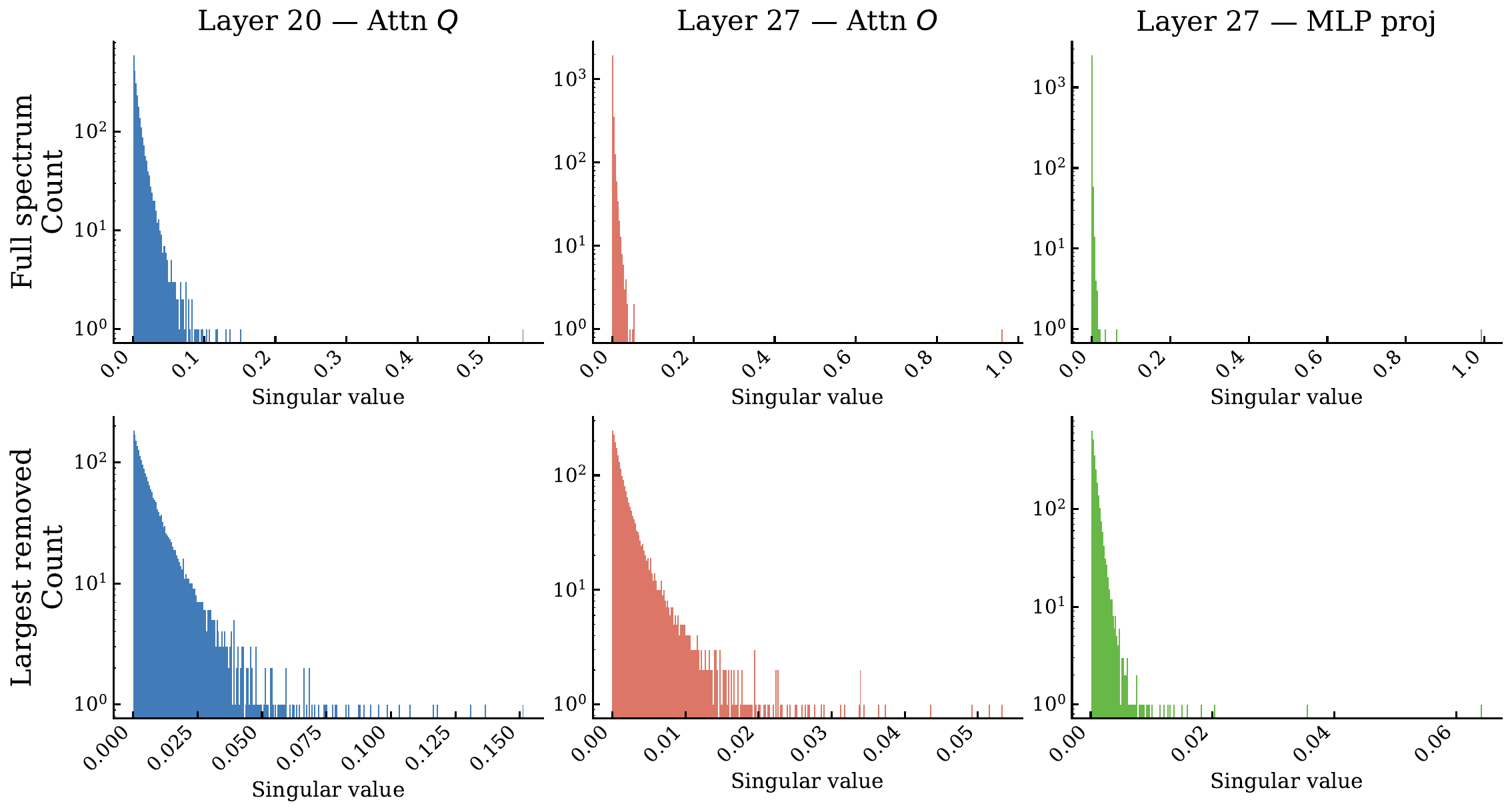}
    \caption{We plot the distribution of normalized singular values
    for the selected momentum matrices of the 2.8B model at step 1500.
    The top row plots the full spectrum, while the bottom row shows
    the same distribution with the leading singular value 
    removed to reveal the bulk.}
    \label{fig:svalue_hist}
\end{figure}

Since all tracked quantiles stabilize, the full spectrum stabilizes
as well. \autoref{fig:svalue_hist} shows the distribution of
normalized singular values for the selected momentum matrices of the
2.8B model at step $1450$. The top row plots the full spectrum, while
the bottom row shows the same distribution with the leading singular
value removed to reveal the bulk. Two consistent features emerge
across all layer types (see
\autoref{fig:svalue_hist_2.8B_1450_all_layers} for the bulk of every
tracked matrix): (i) each spectrum is dominated by a single outlier
singular value, often an order of magnitude or more larger than the
rest of the distribution; and (ii) once this outlier is removed, the
remaining singular values are concentrated near zero, with the count
decaying roughly exponentially as the singular value grows. The scale of the bulk varies markedly across layer types --- a
variation we exploit in \autoref{sec:scaling_laws} when we fit
layer-dependent scaling exponents. This heavy concentration near zero
is precisely what places the late layers at risk of NS failure at
scale.

\subsection{Quantile Dynamics under Rank-$p$ Updates}

\begin{figure}[h]
    \centering
    \includegraphics[width=1.0\textwidth]{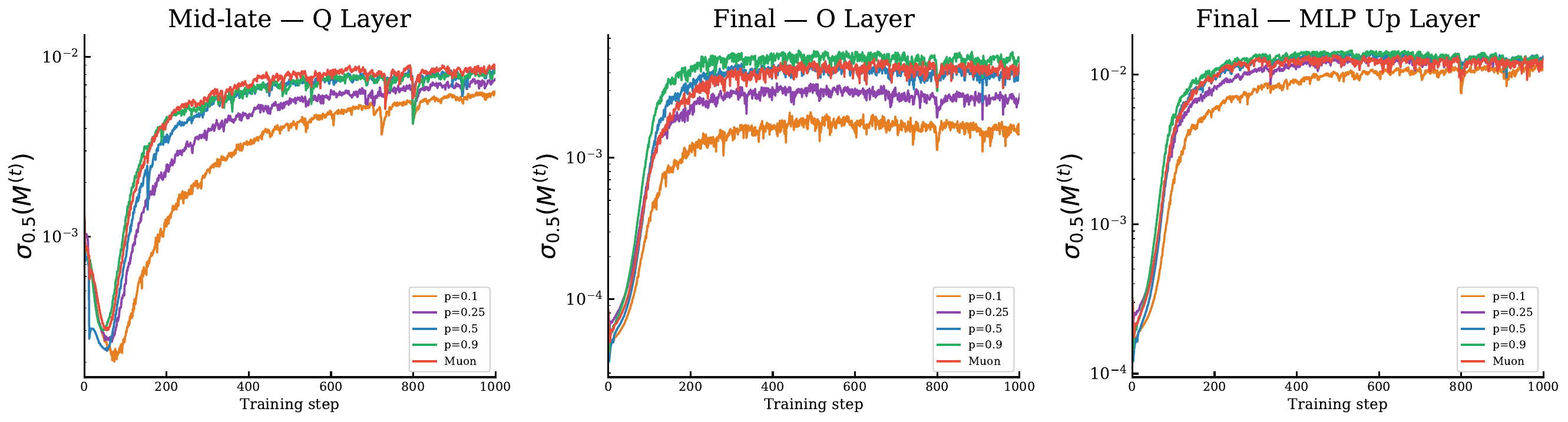}
    \caption{Quantile dynamics for the 50\% quantile of the normalized momentum matrices
    for the 354M model under rank-$p$ orthonormal updates. The trajectories for $p=0.9$ and $p=0.5$
    closely track Muon's quantile, while at $p=0.25$ and $p=0.1$ the deviation grows bigger.}
    \label{fig:quantile_evolution_directions}
\end{figure}

Recall that in \autoref{sec:how_much_normalization_is_needed} we
studied the effect of rank-$p$ orthonormal updates on validation
loss. Here we complement that analysis by examining how the $50\%$
quantile of the normalized momentum matrices evolves under these
updates. As shown in \autoref{fig:quantile_evolution_directions} (for
the 354M model), the trajectories for $p=0.9$ and $p=0.5$ --- the
regimes that closely match Muon's validation loss --- closely track
Muon's quantile as well. At $p=0.25$ the quantile begins to deviate
downward, and at $p=0.1$ the deviation grows further. The same
pattern holds across most tracked layers
(\autoref{fig:quantile_evolution_directions_all_layers}).

This yields a clean correspondence: rank-$p$ updates that closely
track Muon's performance also closely track its singular value
dynamics, while those with degraded performance exhibit deviating,
typically smaller quantile trajectories.

This correspondence has a direct consequence for NS. As long as NS still
orthonormalizes at least the top $50\%$ of directions, 
its induced quantile
dynamics fall under the $p \geq 0.5$ branch above and closely track
full Muon. In this regime the scaling laws we derive in
\autoref{sec:scaling_laws} --- fit to full-Muon stabilization values
--- are \emph{self-consistent}: choosing an NS configuration so that
the predicted $50\%$-quantile sits above its failure threshold will
indeed orthonormalize that fraction at the target scale. For NS
configurations that orthonormalize only the top $25\%$ (or
fewer) of directions at scale, the laws \emph{may} underestimate how
many directions NS misses, since the $p=0.25$ and $p=0.1$
trajectories sit below full Muon's. Quantifying this effect at scale
would require fitting separate laws to rank-$p$ runs, which need a
per-step SVD and are far more expensive than running Muon itself; we
leave this to future work.

\section{Spectral Scaling Laws}
\label{sec:scaling_laws}

As discussed in \autoref{sec:quantiles}, the singular value quantiles
of the momentum matrices stabilize after a short burn-in phase, with
stabilization values that decrease with model size. To predict what
fraction of directions a given NS configuration will orthonormalize
at scale, we need to understand how these stabilization values scale
with model size. We study this scaling for quantiles
$q \in \{0.1, 0.25, 0.5, 0.75, 0.9\}$.

For each model size and layer type, we estimate the stabilization
value by averaging the corresponding quantile over training steps
$1300$--$1500$, and plot it against model size on a log-log scale.
\autoref{fig:scaling_laws} shows the scaling laws for six
representative layer types; we observe a remarkably clean power law
in model size. As shown in \autoref{fig:scaling_laws_quantiles}, the
same pattern holds across all six tracked layer types: for each, all
five quantiles share the same scaling exponent --- and that exponent
depends on the layer type.

The exponents vary substantially across depth. The mid-early, mid,
and mid-late layers scale very mildly with model size, with exponents
around $-0.25$ --- meaning that increasing model size by a factor of
$32$ decreases the stabilization value by only roughly a factor of
$2$. The final layers, in contrast, scale down far more aggressively:
the final MLP projection matrix, for instance, has an exponent of
$-0.96$, so its stabilization value decreases nearly linearly with
model size. Thus, we have a wide range of scaling exponents across
layers, making a uniform NS configuration suboptimal at scale.

\subsection{Case Study: Extrapolating to Frontier Scale}

To illustrate how the fitted laws are used in practice, consider a
$300$B-scale training run (a $\sim\!100\times$ jump from our largest
fitted scale of 2.8B). We compare two contrasting layer types: the
mid-late $Q$ projection and the final $O$ projection. Suppose we
want to orthonormalize at least $50\%$ of the directions in each;
then the relevant quantile is $q=0.5$.
\paragraph{Mid-late $Q$.}
From \autoref{fig:scaling_laws}, the $q=0.5$ stabilization value
at 2.8B is around $5 \cdot 10^{-3}$. The fitted exponent for this
layer type is $-0.27$, so the law predicts a value at 300B of
\[
    5 \cdot 10^{-3} \cdot 100^{-0.27} \;\approx\; 1.4 \cdot 10^{-3}.
\]
This sits above the NanoGPT 5-step failure regime
(\autoref{fig:nanogpt_ns_map}), so the standard 5-step NS
configuration will continue to sufficiently orthonormalize this layer correctly
at 300B.
\paragraph{Final $O$.}
For the final $O$ projection, the $q=0.5$ value at 2.8B is around
$10^{-3}$, with a fitted exponent of $-0.66$. The law predicts a
value at 300B of
\[
    10^{-3} \cdot 100^{-0.66} \;\approx\; 5 \cdot 10^{-5},
\]
which falls inside the NanoGPT failure regime
(\autoref{fig:nanogpt_ns_map}). For this layer one would need a more
accurate NS configuration --- e.g., the 10-step composition used by
DeepSeek-V4 (\autoref{fig:deepseek_v4_ns_map}).
\section{Conclusion}

\label{sec:conclusion}

We presented the first systematic study of how Muon's momentum
spectrum evolves during training and scales with model size. Across
models from 77M to 2.8B parameters and layers at all relative depths,
we identified a consistent picture: after a short burn-in, every
quantile of the momentum spectrum stabilizes at a value determined by
the layer type and model size, and these stabilization values follow
clean power laws in model size with layer-dependent exponents.

The exponents differ markedly across layers --- ranging from roughly
$-0.25$ for mid-early through mid-late layers down to $-0.96$ for the
final MLP projection. This wide range is the central finding of our
paper and has a direct practical consequence: a uniform NS
configuration applied across all layers is unavoidably suboptimal at
scale, since the layers that need the most accurate orthonormalization
are precisely those whose singular values shrink fastest with model
size. Our case study illustrates this concretely: extrapolating from
the 2.8B model, a $300$B-scale training run can continue to use the
$5$-step NanoGPT NS coefficients for the majority of its layers, but
some of the final layers will fall into the NS failure regime unless
a more accurate configuration --- such as the $10$-step composition
used by DeepSeek-V4 --- is applied to those layers.

Together, these results turn a previously opaque design choice ---
how accurate must NS be? --- into a quantitative, layer-aware
decision that can be made directly from our scaling laws. We see
several natural extensions. First, the stabilization phenomenon and
the particular exponents we measure may be specific to GPT-2-style
language models trained with Muon. Studying analogous scaling laws
for other architectures (e.g., Mixture-of-Experts models) and for
other optimizers that rely on iterative matrix-function
approximations --- most notably
Shampoo~\citep{gupta2018shampoo} and its
descendants~\citep{vyas2024soap, eschenhagen2025purifying} --- is a
natural next step. Second, designing NS coefficients specifically
tuned to the empirical singular value distribution of each layer is
a promising avenue for further reducing the cost of orthonormalization
at frontier scale. We leave both directions to future work.

\newpage

\bibliographystyle{plainnat}
\bibliography{gmref}


\appendix

\section{Appendix}

\subsection{Details on pre-training}
\label{sec:pre_training_details}

We used the modded-nanogpt codebase \citep{modded_nanogpt_2024} for all experiments.
All matrix-valued parameters are trained with Muon, while non-matrix parameters
(embeddings, LM head, and biases) are trained with AdamW with
$(\beta_1, \beta_2) = (0.9, 0.95)$ and learning rate $0.002$. For both optimizers we
use a weight decay of $0.01$ throughout. For clarity, initialization scaling is
omitted from Algorithm \autoref{alg:muon}; in practice, we scale matrix parameters by
$\sqrt{d_{\text{out}}/d_{\text{in}}}$ and LM head parameters by $1/\sqrt{d_{\text{in}}}$,
which promotes learning rate transfer across scales \citep{pethick2025training}.

For the experiments in \autoref{sec:how_much_normalization_is_needed}, we tuned the
learning rate for 77M models over the grid $\{0.01, 0.02, 0.03, 0.04, 0.05\}$.
We observed little sensitivity and found $0.03$ to be optimal across all rank-$p$
configurations; we adopted it for the 160M and 354M models as well. For the
experiments in \autoref{sec:quantiles}, we observed no meaningful difference
between learning rates $0.01$, $0.02$, and $0.03$ at the 160M scale; we therefore
fixed the learning rate to $0.01$ across all model sizes, leveraging Muon's known
learning-rate transfer property. We used a constant learning rate followed by
linear decay over the final 10\% of training. All models are trained on the
FineWeb dataset \citep{penedo2024fineweb} for a Chinchilla-optimal token budget
\citep{hoffmann2022chinchilla} of $20\times$ the number of model parameters. All
experiments were run on L40 or H200 GPUs, with larger models requiring 2 GPUs.

\subsection{On the Newton Schultz Approximation}
\label{sec:ns_approximation}

\subsubsection{NanoGPT NS Coefficients}

Here we present the NS polynomials used and popularized by the 
NanoGPT speedrun \citep{modded_nanogpt_2024, cesista2025muonoptcoeffs}.

\begin{equation}
\begin{aligned}
    b_1(x) &= 4.0848x - 6.8946x^3 + 2.9270x^5 \\
    b_2(x) &= 3.9505x - 6.3029x^3 + 2.6377x^5 \\
    b_3(x) &= 3.7418x - 5.5913x^3 + 2.3037x^5 \\
    b_4(x) &= 2.8769x - 3.1427x^3 + 1.2046x^5 \\
    b_5(x) &= 2.8366x - 3.0525x^3 + 1.2012x^5
\end{aligned}
\label{eq:nanogpt_ns_coefficients}
\end{equation}

The full NS map is then $f = b_5 \circ b_4 \circ b_3 \circ b_2 \circ b_1$.

\begin{figure}[h]
    \includegraphics[width=0.8\textwidth]{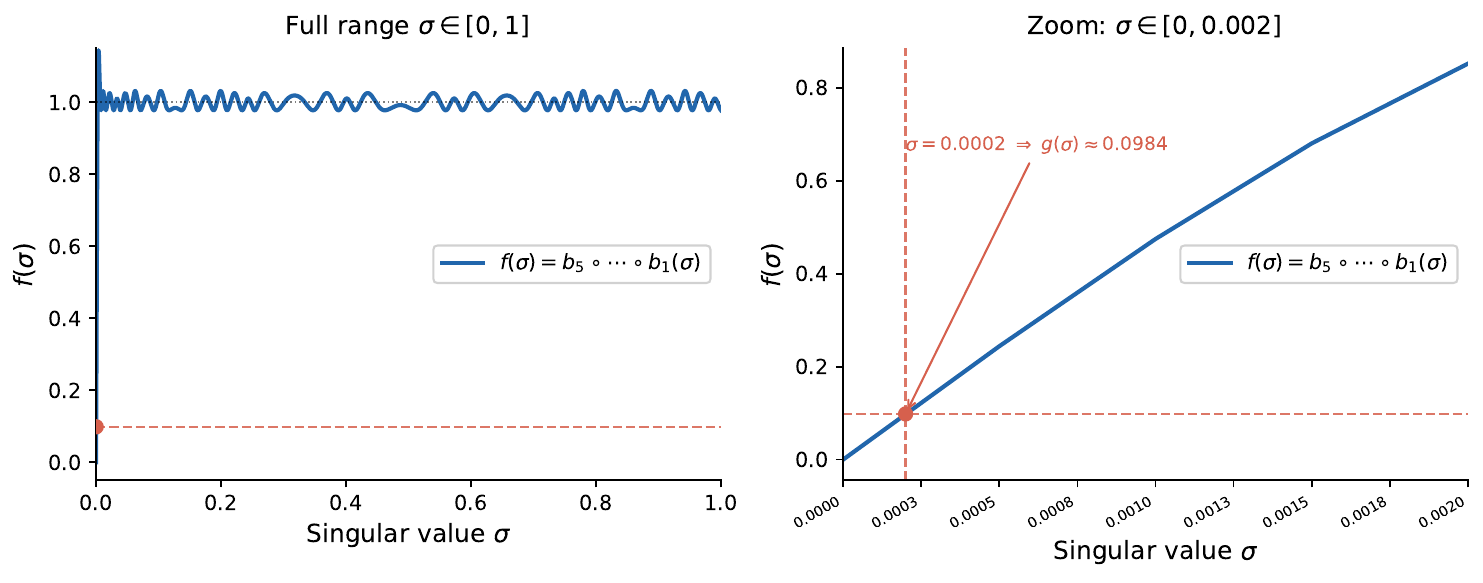}
    \caption{The NS map $f(\sigma) = b_5 \circ b_4 \circ b_3 \circ b_2 \circ b_1(\sigma)$ for $b_i$ in \autoref{eq:nanogpt_ns_coefficients}.}
    \label{fig:nanogpt_ns_map}
\end{figure}

\subsubsection{DeepSeek-V4 NS Coefficients}

\citep{deepseekv4} uses the following NS coefficients:

\begin{equation}
\begin{aligned}
&a(x) = 2x - 1.5x^3 + 0.5x^5 \\
&c(x) = 3.4445x - 4.7750x^3 + 2.0315x^5 \\
\end{aligned}
\label{eq:deepseek_v4_ns_coefficients}
\end{equation}

\begin{figure}[h]
    \includegraphics[width=0.8\textwidth]{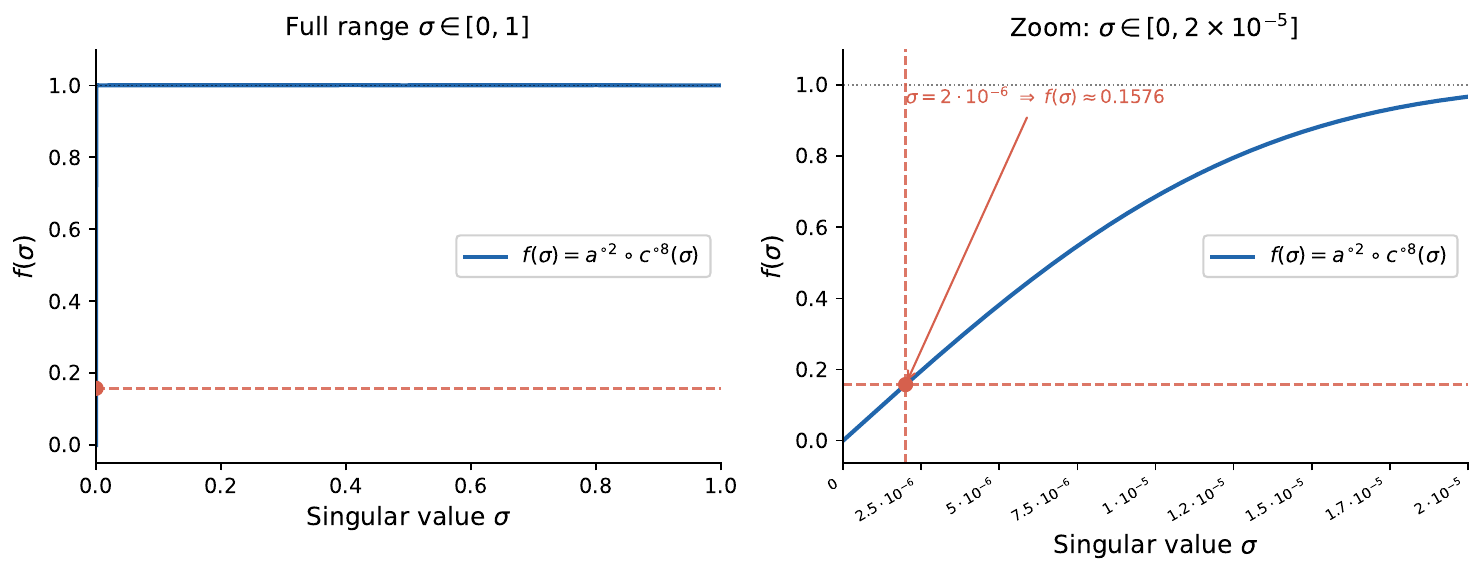}
    \caption{The NS map $f(\sigma) = a^{\circ 2} \circ c^{\circ 8}(\sigma)$ for 
    $a$ and $c$ in \autoref{eq:deepseek_v4_ns_coefficients}.}
    \label{fig:deepseek_v4_ns_map}
\end{figure}

The full NS map is then $f = a^{\circ 2} \circ c^{\circ 8}$.
While this approximation is very good (see \autoref{fig:deepseek_v4_ns_map}), 
is uses 10 steps and hence is computationally more expensive.

\newpage
\section{Appendix B}

\begin{figure}[h]
    \centering
    \begin{minipage}{0.48\textwidth}
        \centering
        \includegraphics[width=\textwidth]{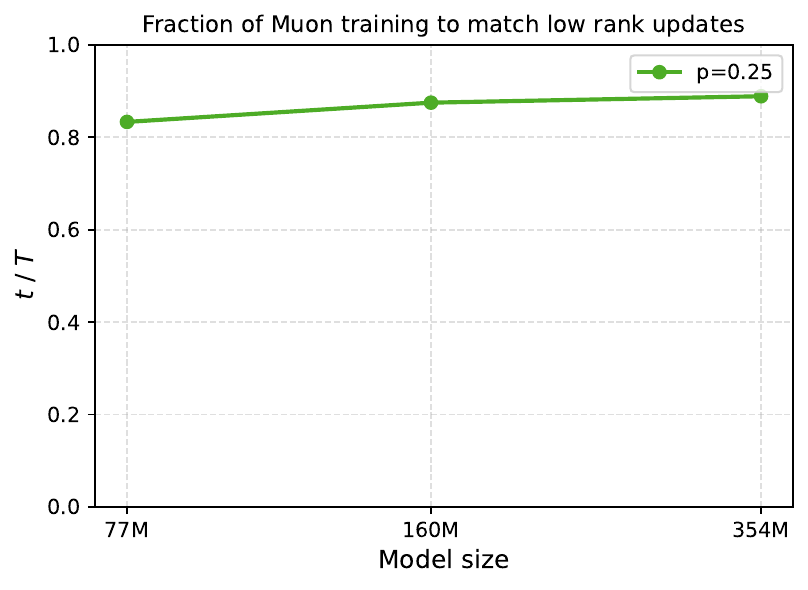}
        \caption{We observe that Muon needs around $(80-90)\%$ of the iterations to 
        match the final loss of the rank $p=0.25$ run.}
        \label{fig:ablation_directions_matching_25}
    \end{minipage}
    \hfill
    \begin{minipage}{0.48\textwidth}
        \centering
        \includegraphics[width=\textwidth]{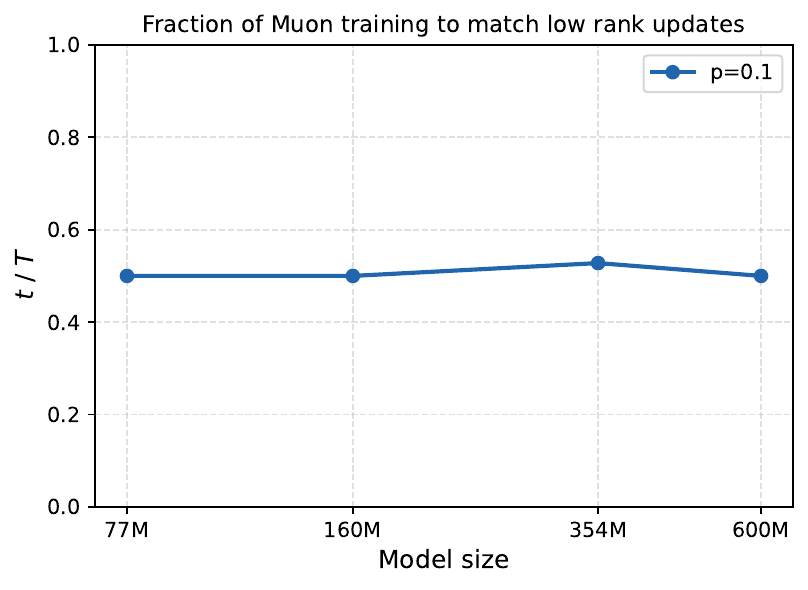}
        \caption{We observe that Muon needs around $(50-55)\%$ of the iterations to 
        match the final loss of the rank $p=0.1$ run.}
        \label{fig:ablation_directions_matching_10}
    \end{minipage}
\end{figure}

\begin{figure}[h]
    \centering
    \includegraphics[width=1.0\textwidth]{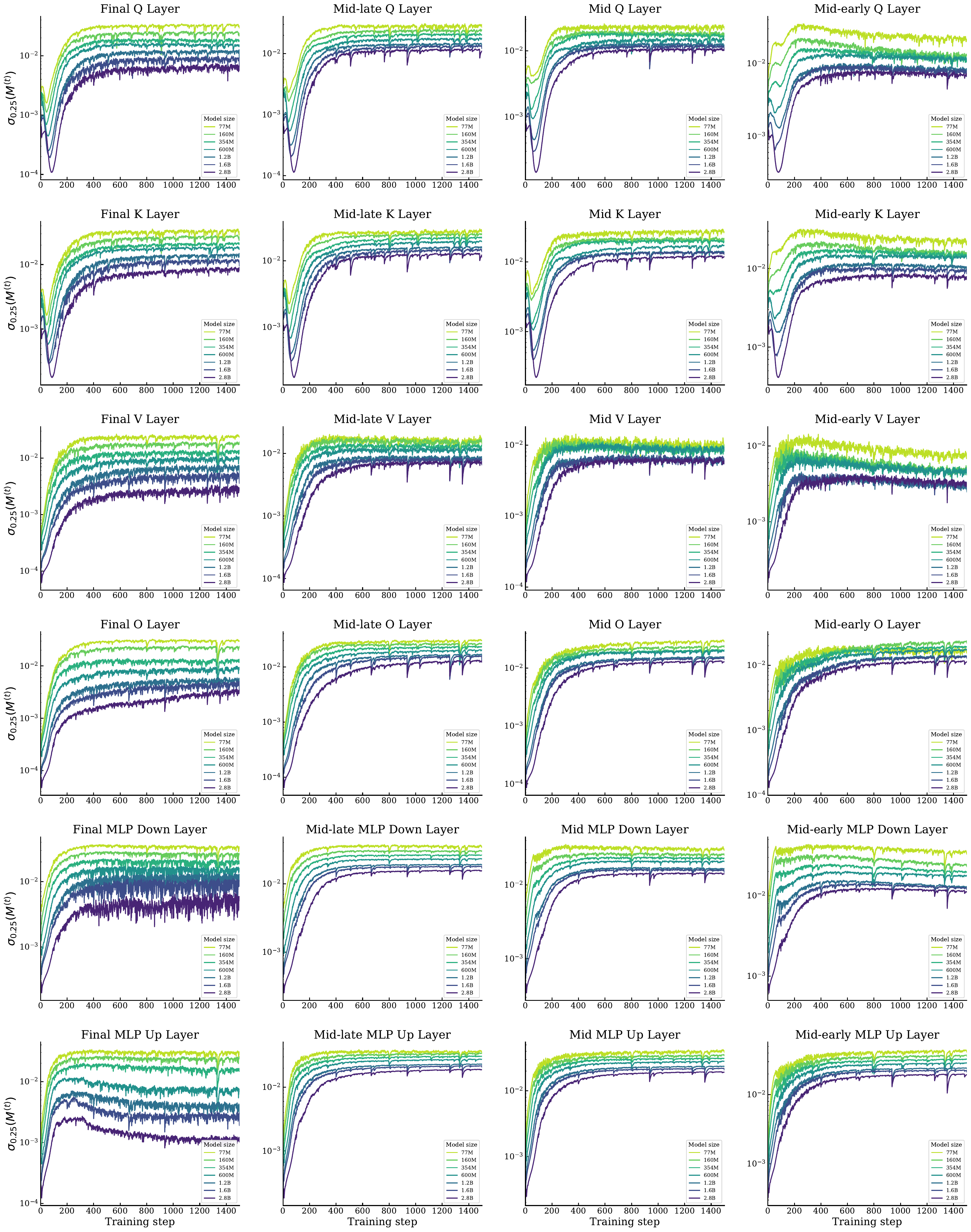}
    \caption{Quantile evolution for the 25\% quantile for all layer types and model sizes.}
    \label{fig:quantile_evolution_25}
\end{figure}

\begin{figure}[h]
    \centering
    \includegraphics[width=1.0\textwidth]{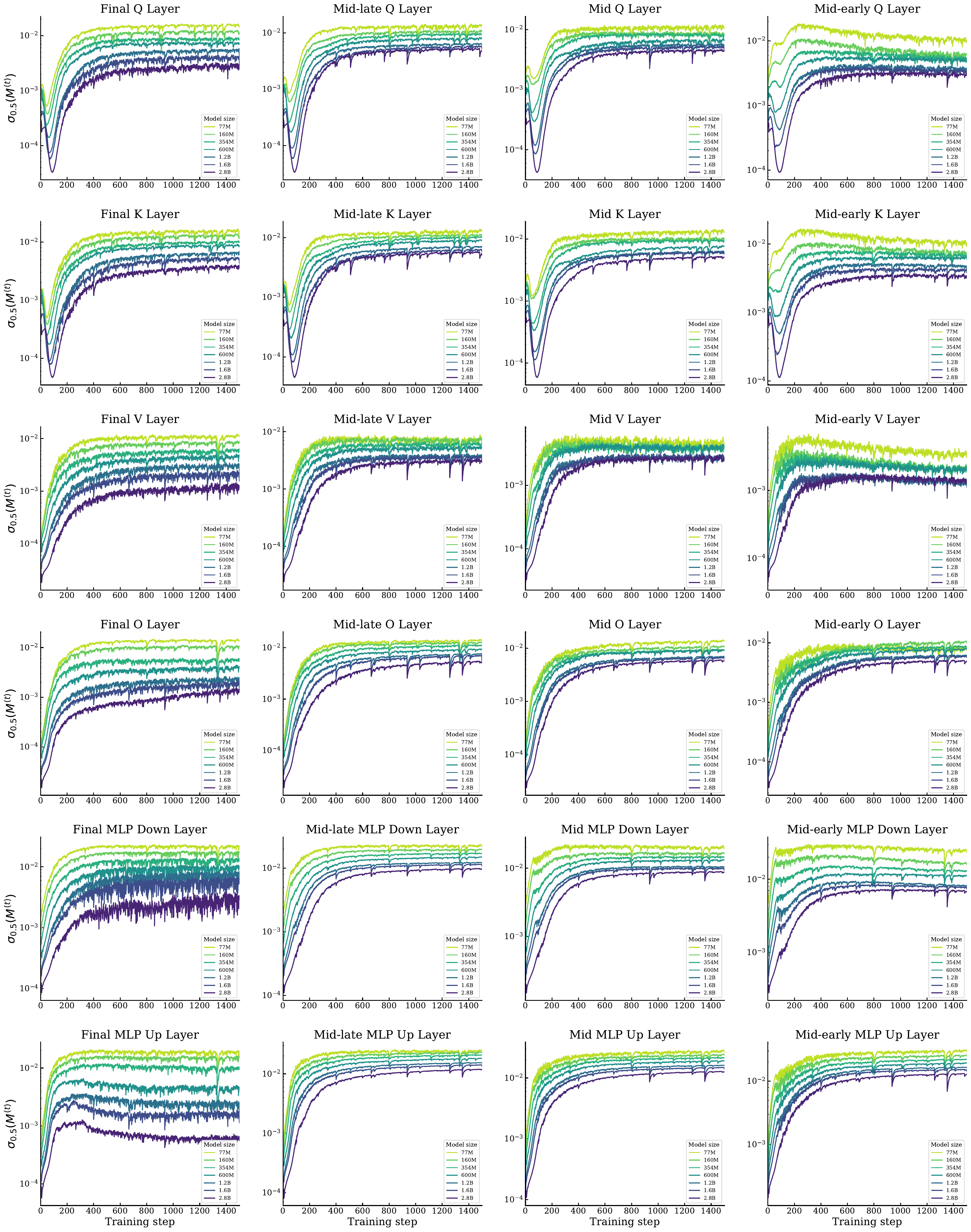}
    \caption{Quantile evolution for the 50\% quantile for all layer types and model sizes.}
    \label{fig:quantile_evolution_10}
\end{figure}

\begin{figure}[h]
    \centering
    \includegraphics[width=1.0\textwidth]{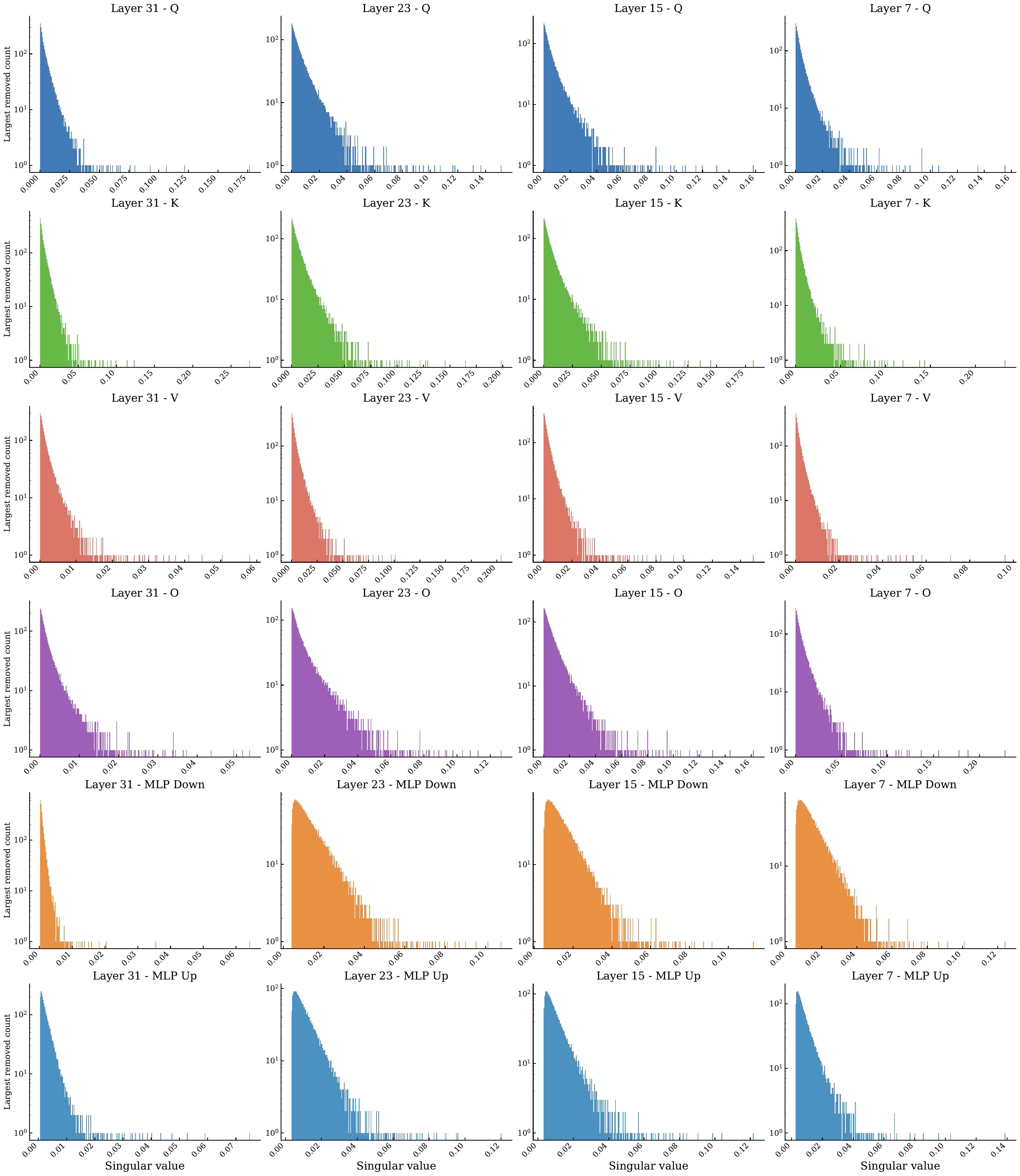}
    \caption{Normalized singular value spectra of the 2.8B model at step 1450, with the dominant singular value removed, shown for every tracked weight matrix.}
    \label{fig:svalue_hist_2.8B_1450_all_layers}
\end{figure}

\begin{figure}[h]
    \centering
    \includegraphics[width=1.0\textwidth]{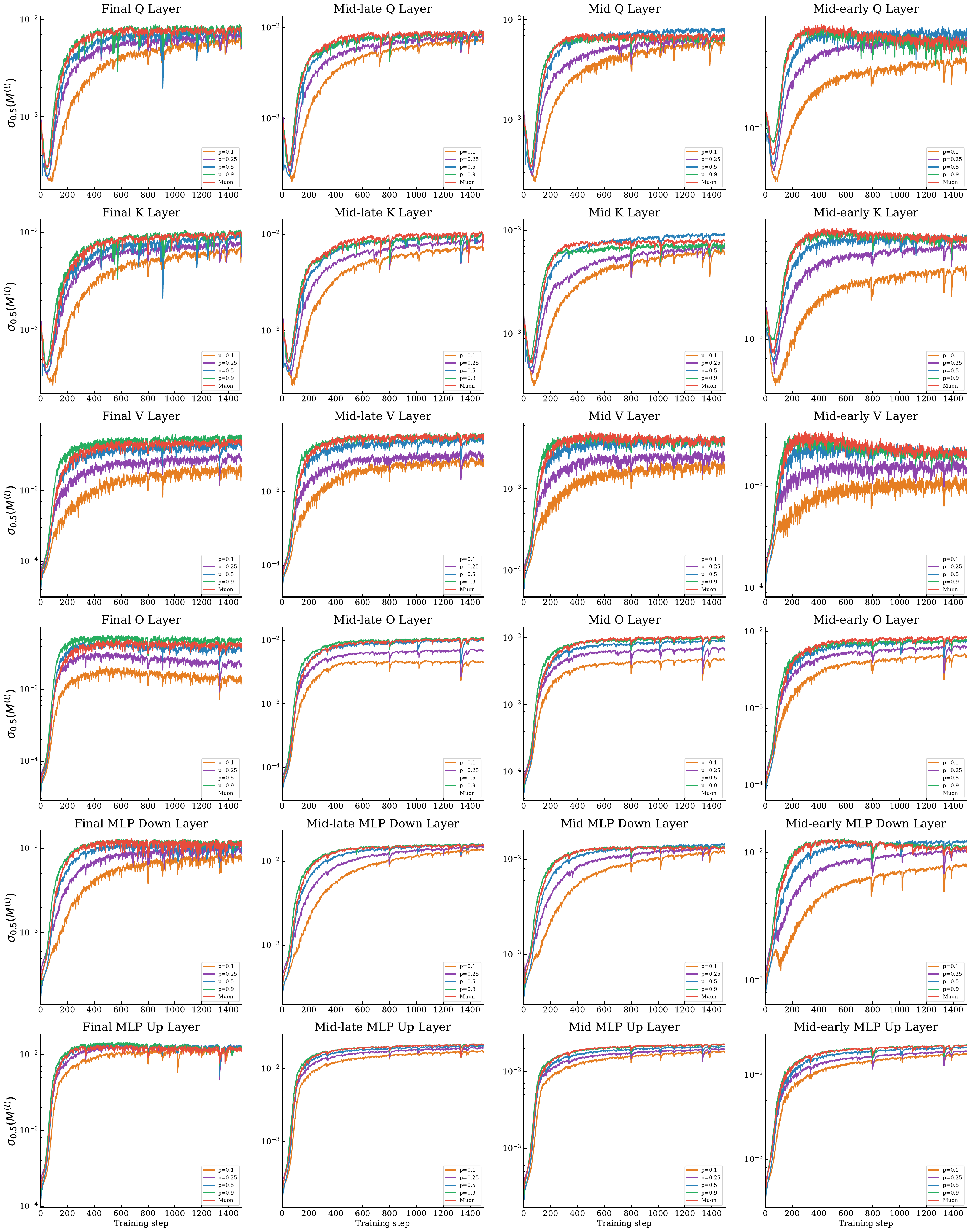}
    \caption{Quantile dynamics for the 50\% quantile of the normalized momentum matrices
    for the 354M model under rank-$p$ orthonormal updates. The trajectories for $p=0.9$ and $p=0.5$
    closely track Muon's quantile, while at $p=0.25$ and $p=0.1$ the deviation grows bigger.}
    \label{fig:quantile_evolution_directions_all_layers}
\end{figure}

\begin{figure}[h]
    \centering
    \includegraphics[width=1.0\textwidth]{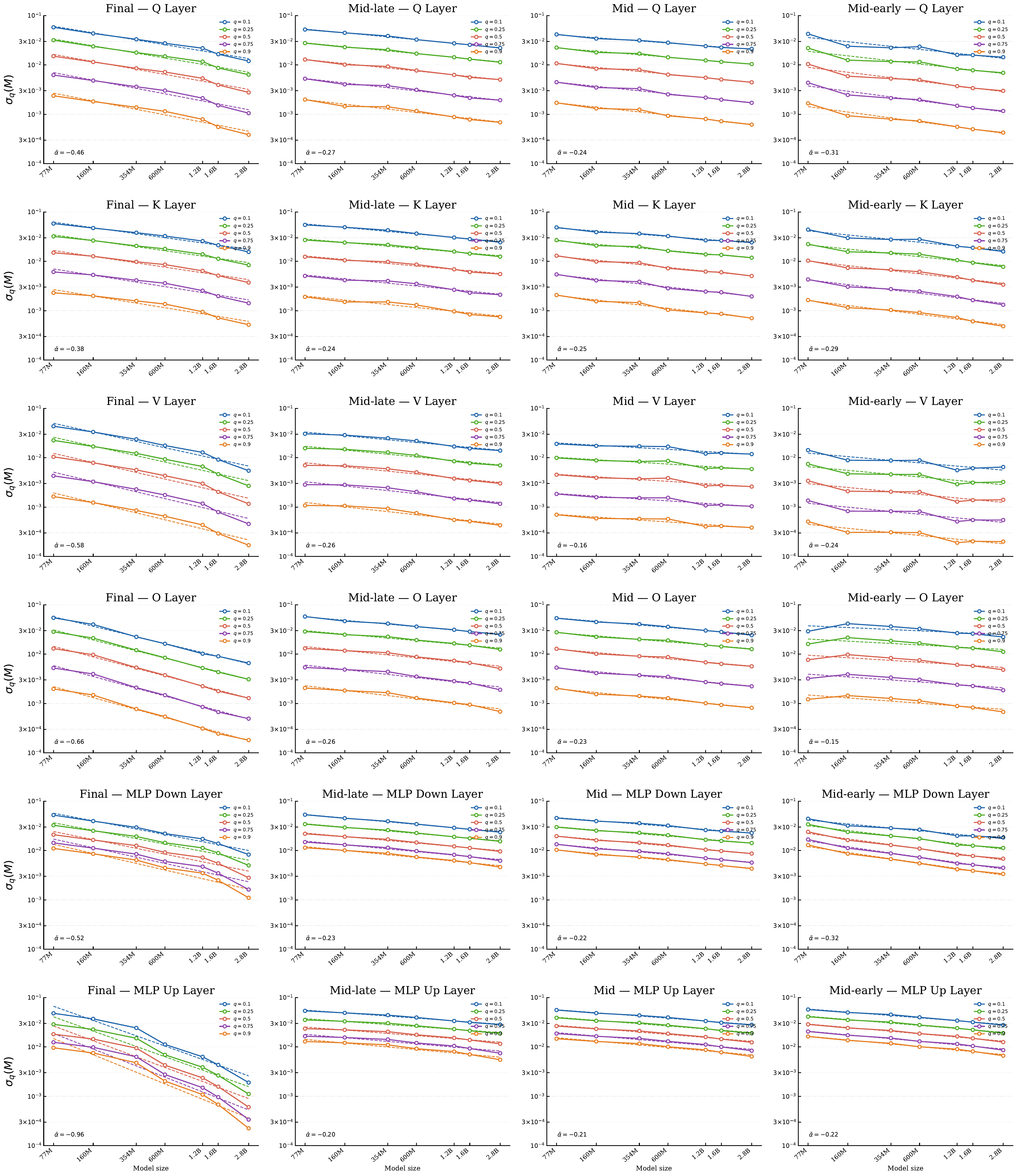}
    \caption{Scaling laws for all tracked quantiles and layer types across model sizes.}
    \label{fig:scaling_laws_quantiles}
\end{figure}

\newpage

\end{document}